%% file: main.tex
\documentclass{svproc}
\usepackage{url}
\usepackage{wrapfig}
\usepackage{graphicx}
\usepackage{xcolor}
\usepackage{tikz}
\usepackage{amsmath,amsfonts,mathtools,amssymb}
\usepackage{pgfplots}
\usepgfplotslibrary{groupplots}
\usepackage{siunitx}
\usepackage{algorithm}
\usepackage[noend]{algpseudocode}
\usepackage{nicefrac}
\usepackage{yhmath}
\usepackage{comment}
\usepackage{hyperref}
\hypersetup{colorlinks,allcolors=black}
\usepackage{subcaption} 
\usepackage{aligned-overset}
\usepackage[style=numeric-comp, backend=biber,sortcites = true,maxbibnames = 99, maxcitenames = 99, mincitenames = 2, giveninits=true, sorting=none]{biblatex}
\input{macros}

\usepackage{fancyhdr}
\newcommand{\mytitle}{\textbf{Accepted version.}
Accepted to the Symposium on Systems Theory in Data and Optimization (SysDO 2024).
This is a preprint of the final version, which is to appear in Lecture Notes in Control and Information Sciences -
Proceedings.
}
\begin{document}
\mainmatter              
\title{{PACSBO}: Probably approximately correct safe Bayesian optimization}
\author{Abdullah Tokmak\inst{1} \and Thomas B.\ Schön\inst{2} \and
Dominik Baumann\inst{1,2}}
\authorrunning{Tokmak et al.}
\tocauthor{Abdullah Tokmak, Thomas B.\ Schön, Dominik Baumann}
\institute{Cyber-Physical Systems Group, Aalto University, Espoo, Finland \\
\email{firstname.lastname@aalto.fi}\\ 
\and
Department of Information Technology, Uppsala University, Uppsala, Sweden \\
\email{thomas.schon@it.uu.se}
}


\maketitle              

\thispagestyle{fancy}	
\pagestyle{empty}

\begin{abstract}
Safe Bayesian optimization (BO) algorithms promise to find optimal control policies  without knowing the system dynamics while at the same time guaranteeing safety with high probability. 
In exchange for those guarantees, popular algorithms require a smoothness assumption: a known upper bound on a norm in a reproducing kernel Hilbert space (RKHS).
The RKHS is a potentially infinite-dimensional space, and it is unclear how to---in practice---obtain an upper bound of an unknown function in its corresponding RKHS. 
In response, we propose an algorithm that estimates an upper bound on the RKHS norm of an unknown function from data and investigate its theoretical properties.
Moreover, akin to Lipschitz-based methods, we treat the RKHS norm as a local rather than a global object, and thus reduce conservatism. 
Integrating the RKHS norm estimation and the local interpretation of the RKHS norm into a safe BO algorithm yields \pacsbo, an algorithm for probably approximately correct safe Bayesian optimization, for which we provide numerical and hardware experiments that demonstrate its applicability and benefits over popular safe BO algorithms.
\end{abstract}

\input{Sections/introduction}
\input{Sections/preliminaries}
\input{Sections/algorithm}
\input{Sections/approximation}

\input{Sections/experiments}

\input{Sections/summary}

\input{Sections/appendix}
\section*{Acknowledgements}
The authors thank Sara Pérez and Mingwei Deng for helpful comments.
\printbibliography

\end{document}

%% file: macros.tex
\newtheorem{assumption}{\bf Assumption}

\newcommand{\capt}[1]{\mdseries{\emph{#1}}}

\newcommand{\safeopt}{\textsc{SafeOpt}}
\newcommand{\pacsbo}{\textsc{Pacsbo}}
\newcommand{\gpucb}{\textsc{GP-UCB}}

\newcommand{\ie}{i\/.\/e\/.,\/~}
\newcommand{\eg}{e\/.\/g\/.,\/~}

\newcommand{\domain}{\mathcal{A}}
\newcommand{\Ig}{\mathcal{I}_\mathrm{g}}

\newcommand{\fakepar}[1]{\vspace{1mm}\noindent\textbf{#1.}}

\newcommand*\diff{\mathop{}\!\mathrm{d}}

\definecolor{aaltoBlue}{RGB}{0,94,184}%


%% file: Sections/introduction.tex

\section{Introduction}\label{sec:intro}
In recent years, reinforcement learning (RL) has achieved remarkable success in controlling high-dimensional systems without requiring a dynamics model~\cite{Duan2016Benchmarking, Lillicrap2019Continuous}.
However, most of the impressive results have been obtained in simulation environments.
Applying RL algorithms to real-world robotic systems is challenging because
\emph{(i)}~when interacting with real-world environments, it is crucial to guarantee safety, and
\emph{(ii)} each sample corresponds to a potentially expensive experiment, and hence sample efficiency is essential.
Popular RL algorithms fail to provide both safety and sample efficiency.

Combining Gaussian process (GP) regression~\cite{Rasmussen2006Gaussian} with Bayesian optimization (BO)~\cite{Frazier2018Bayesian} provides a sample efficient alternative to RL and has been successfully applied on various hardware platforms~\cite{antonova2017deep,calandra2016bayesian,marco2016automatic}.
Based on GP regression and BO, several algorithms have been proposed that can, in addition, provide probabilistic safety guarantees~\cite{Sui2015Safe,Berkenkamp2021Bayesian}.
These safe learning algorithms aim at optimizing an unknown reward function while satisfying (unknown) constraints with high probability.
In exchange for the safety guarantees, they require smoothness assumptions for reward and constraint functions.
Particularly, they assume that reward and constraint functions have a known upper bound in a reproducing kernel Hilbert space (RKHS).
The RKHS norm is a norm in a potentially infinite dimensional space, and assuming exact knowledge of that norm (or a tight upper bound on it) in unknown environments is highly unrealistic.
Notably, a loose upper bound on the RKHS norm leads to conservative algorithms, whereas an underestimation might lead to constraint violations.
We address this shortcoming by estimating the RKHS norm from data and incorporating the estimate into a safe BO algorithm.

Moreover, popular BO algorithms treat the RKHS norm as a global object~\cite{Sui2015Safe,Berkenkamp2021Bayesian,Srinivas2010Gaussian}.
Since functions may exhibit vastly different smoothness properties on different parts of their parameter space, Lipschitz-based methods regularly consider a local Lipschitz constant to reduce conservatism~\cite{Calliess2020Lazy}.
Inspired by Lipschitz-type approaches, we introduce a local interpretation of the RKHS norm.

\fakepar{Related Work}
Learning control policies while providing safety guarantees has attracted growing interest over the past years; see~\cite {Brunke2022Review} for a review. In the area of GP regression and BO, \safeopt~\cite{Sui2015Safe,Berkenkamp2021Bayesian}, which is based on \gpucb-type algorithms~\cite{Srinivas2010Gaussian, Chowdhury2017Kernelized}, is arguably the most well-known and popular safe learning algorithm.
Moreover, various algorithms based on \safeopt\ have been proposed, targeting, \eg better scalability~\cite{Sui2018Stage, duivenvoorden2017constrained} or global optimization~\cite{Bhavi2023GSO}.
However, all aforementioned works require a known, tight upper bound on the RKHS norm, and it is unclear how to compute a reliable RKHS norm over-estimation of an unknown function.
The recent publication~\cite{fiedler2024safety} acknowledges this downside of popular BO algorithms and proposes an alternative approach requiring instead the Lipschitz constant and an upper bound on the noise amplitude, both of which are usually unknown.
Especially estimating the Lipschitz constant of unknown functions is a delicate problem in itself~\cite{Wood1996Lipschitz}.

An alternative to assuming a known RKHS norm upper bound is estimating it from data, which has not been extensively investigated. References~\cite{Hashimoto2020Nonlinear, Scharnhorst2022Robust} discuss ways to under-estimate the unknown RKHS norm.
Nevertheless, relying on an underestimation of the RKHS norm implies overly optimistic exploration, which can lead to constraint violations.
Also,~\cite{Tokmak2023Alkiax} introduces a simple heuristic RKHS norm extrapolation to estimate an upper bound on the RKHS norm, however, without providing any guarantees.
We draw inspiration from the ideas in~\cite{Scharnhorst2022Robust,Hashimoto2020Nonlinear, Tokmak2023Alkiax} and propose an algorithm that estimates an upper bound on the RKHS norm from data, eliminating the need for correctly guessing the RKHS norm. 
Moreover, we investigate the theoretical properties of that algorithm.

\fakepar{Contribution}
In this paper, we propose \pacsbo, a novel algorithm for probably approximately correct safe BO.
We make the following contributions:
\begin{enumerate}
    \item We propose an algorithm that estimates the RKHS norm of an unknown function with probabilistic guarantees.
    Hence, we eliminate the need to guess the RKHS norms of the reward and constraint functions correctly.
    \item We treat the RKHS norm as a local object. As a result, we can explore less conservatively.
    \item We extend \safeopt\ with the RKHS norm over-estimation and the local interpretation of the RKHS norm, yielding \pacsbo.
    \item We show the benefits of \pacsbo\ compared to \safeopt\ in numerical experiments. Furthermore, we demonstrate the successful deployment of \pacsbo\ on a real Furuta pendulum.
\end{enumerate}

%% file: Sections/preliminaries.tex
\section{Problem setting and preliminaries} \label{section:p_and_p}

We propose a safe BO algorithm that does not require a priori knowledge of the RKHS norm.
In the following, we make the problem setting precise and introduce preliminaries, in particular, \safeopt.

\fakepar{Problem setting}
We formulate the learning problem as optimizing an unknown reward function $f{:}\;\domain\to\mathbb{R}$ that maps from the parameter space $\domain\subseteq \mathbb{R}^n$ with parameters $a\in\domain$ to a scalar reward value, while guaranteeing safety.
We express safety through unknown constraint functions $g_i{:}\;\domain\to \mathbb{R}, i\in \Ig=\{1,\ldots, s\}$.
Akin to~\cite{Berkenkamp2021Bayesian}, we introduce the surrogate function~$h{:}\; \domain \times \mathbb{N} \rightarrow \mathbb R$ with~$h(a,i)=f(a)$ if~$i=0$, and $h(a,i)=g_i(a)$ if~$i \in \Ig$, and define~$\mathcal I \coloneqq \{0\} \cup \Ig$.
Hence, we can write the optimization problem as
\begin{align} \label{eq:opt}
    \max_{a \in \domain} h(a,0) \quad \text{subject to } \; h(a,i) \geq 0, \quad \forall i \in \Ig.
\end{align}
In practice, the reward function~$h(\cdot,0)$ may correspond to the control performance, the parameters~$a$ to the parameters of the control policy, and the constraints~$h(\cdot,i), i \in \Ig$, to the distance to obstacles. 

\fakepar{\safeopt}
As~$h$ is unknown, solving~\eqref{eq:opt} is infeasible without further assumptions. 
Thus, we make three assumptions analogous to \safeopt.
\emph{(i)}, we assume that we can conduct experiments with parameters~$a^\prime$ and, in return, receive measurements $\hat{h}(a^\prime,i)=h(a^\prime, i)+\epsilon$, where~$\epsilon$ is $\sigma$-sub-Gaussian measurement noise.
We define the set of parameters that we have already tried as~$A\coloneqq[a_1, \ldots, a_{\mathrm N}]^\top$, and the corresponding measurements as~$\hat h_{A,i}\coloneqq [\hat h(a_1,i),\ldots, \hat h(a_{\mathrm N},i)]^\top$ for all~$i \in \mathcal I$.
\emph{(ii)}, we assume that a set~$S_0$ with~$\emptyset\neq S_0\subseteq\domain$ of initial safe parameters is given, and \emph{(iii)} that~$h(\cdot, i)$ is a member of the RKHS of kernel~$k$, \ie $h(\cdot, i)\in H_k$ for all~$i\in\mathcal I$.
This allows us to provide safety guarantees when estimating $h$ with GPs if, in addition, an upper bound on the RKHS norm is available~\cite[Theorem~4.1]{Berkenkamp2021Bayesian}.
The GP posterior mean and covariance are 
\begin{equation} \label{eq:GP}
\begin{aligned}
        \mu_A(a,i) &= k_A(a)^\top(K_A+\sigma^2 I_N)^{-1}\hat h_{A,i} \\
    \sigma_A^2(a) &= k(a,a) - k_A(a)^\top(K_A+\sigma^2 I_N)^{-1}k_A(a),
\end{aligned}
\end{equation}
respectively~\cite{Rasmussen2006Gaussian}.
We denote the kernel evaluated at~$a\in\domain$ by~$k(a,a)$, the covariance vector by~$k_A(a)=[k(a,a_1), \ldots, k(a,a_{\mathrm{N}})]^\top\in\mathbb{R}^N$, the covariance matrix with entry~$k(a_i,a_j)$ at row~$i$ and column~$j$ for all~$i,j\in\{1,\ldots,\mathrm{N}\}$ by~$K_A\in\mathbb{R}^{N\times N}$, and the identity matrix by~$I_N\in\mathbb{R}^{N\times N}$.
We can now obtain confidence intervals around the posterior mean that contain the ground truth~$h$ with high probability~\cite[Theorem~1]{Chowdhury2017Kernelized}.
With these confidence intervals, we can probabilistically quantify whether a policy parameter~$a\in\domain$ is safe.
Particularly, we restrict function evaluations to a safe set~$S_A\subseteq \domain$ that only contains parameters~$a$ that are safe with high probability, \ie parameters that correspond to constraint values greater than zero.
We start off with~$S_A=S_0$ and sequentially augment the GP model by conducting experiments, leading to a monotonically growing safe set~$S_A$.
However, not all~$a\in S_A$ are evaluated.
In particular, we compute a set of potential maximizers~$M_A\subseteq S_A$ and a set of potential expanders~$G_A\subseteq S_A$ and choose~$a=\arg\max_{M_A \cup G_A} \sigma_A(a)$ at each iteration, thus, balancing exploration and exploitation while guaranteeing safety.
The acquisition function and the definitions of the sets~$S_A,M_A,$ and~$G_A$ are equivalent to \safeopt.
Therefore, we refer to the computation of these sets as the \safeopt-subroutine and point to~\cite{Sui2015Safe} for a detailed introduction.

\begin{wrapfigure}[11]{r}{0.5\textwidth}
\centering
\vspace*{-30pt}
\input{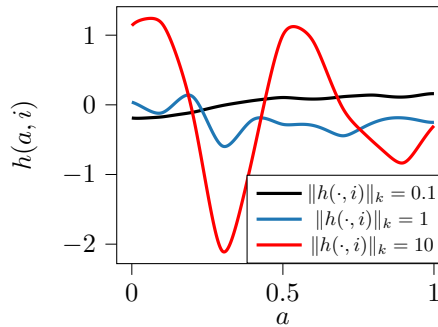}
\vspace*{-23pt}
    \caption{Different RKHS norms.}
    \label{fig:different_RKHS_norms}
\end{wrapfigure}
\fakepar{RKHS norm}
The confidence intervals and the resulting safety and exploration guarantees of \safeopt~\cite[Theorem~4.1]{Berkenkamp2021Bayesian} depend on~$B_i$, an upper bound on the unknown RKHS norm~$\|h(\cdot, i)\|_k$.
The RKHS norm of the function~$h(\cdot, i)\in H_k$ is given by
$
    \|h(\cdot, i)\|_k^2 = \sum_{s=1}^\infty \sum_{t=1}^\infty \alpha_s \alpha_tk(x_s,x_t)
$
with~$\alpha$ the coefficients and~$x$ the center points, which can both be infinite-dimensional~\cite[Section~4]{Steinwart2008SVM}.
Figure~\ref{fig:different_RKHS_norms} illustrates the smoothness of functions given their RKHS norm. 
As mentioned in Section~\ref{sec:intro}, it is unclear how to get a tight upper bound on the RKHS norm of unknown functions.
Clearly, if we assume a higher RKHS norm bound, we may be overly conservative, while assuming a lower RKHS norm bound may result in unsafe explorations.
In this work, we drop the assumption of knowing a tight upper bound~$B_i$. 
Instead, we propose an algorithm that estimates the RKHS norm from data while assuming~$h(\cdot, i)\in H_k$, which is significantly less restrictive.
We investigate the theoretical properties of this algorithm and improve the exploration behavior by treating the RKHS norm as a \emph{local} object to, \eg exploit local smoothness and thus explore less conservatively.

%% file: Sections/algorithm.tex
\section{Proposed algorithm} \label{sec:pacsbo}
In this section, we develop \pacsbo, a novel algorithm for probably approximately correct safe BO. This includes the estimation of an upper bound for the RKHS norm of an unknown function, the treatment of the RKHS norm as a local object, and a theoretical exposition.

\subsection{RKHS norm estimation}\label{sec:RKHS_norm_approx}

\begin{wrapfigure}[10]{r}{0.4\textwidth}
\centering
\vspace*{-45pt}
\input{final_figs/RKHS_norm_variance}
    \vspace*{-25pt}
    \caption{RKHS norm over $r_{\domain,A}$.}
    \label{fig:RKHS_variance}
\end{wrapfigure}
As introduced in Section~\ref{section:p_and_p}, given samples~$A\subseteq\domain$, we estimate~$h(\cdot, i)$ using GPs with mean~$\mu_A(\cdot, i)$ and covariance~$\sigma_A(\cdot)^2$.
For an increasing set~$A$ sampled unbiasedly over~$\domain$, the mean~$\mu_A(\cdot,i)$ serves as a close representation of the ground truth~$h(\cdot,i)$ on~$\domain$.
Thus, the RKHS norm~$\|\mu_A(\cdot,i)\|_k$ is a reasonable estimate of~$\|h(\cdot,i)\|_k$ given a sufficiently high sampling density.
%
%
To quantify the sampling density, we introduce the reciprocal covariance integral~$r_{\domain,A}\coloneqq (\int_{a\in \domain} \sigma_A(a)^2 \diff a)^{-1}$.
Intuitively, as we collect more samples over~$\domain$, the covariance shrinks, and hence its reciprocal integral increases. 
Thus, for an increasing~$r_{\domain,A}$, we would expect~$\|\mu_A(\cdot, i)\|_k$ to approach~$\|h(\cdot, i)\|_k$.
This expectation is empirically confirmed by Figure~\ref{fig:RKHS_variance}.
As~$r_{\domain,A}$ increases, we see that~$\|\mu_A(\cdot, i)\|_k$ approaches~$\|h(\cdot, i)\|_k$ from \emph{below}, as observed in~\cite{Hashimoto2020Nonlinear, Scharnhorst2022Robust}.
Thus, directly using~$\|\mu_A(\cdot,i)\|_k$ as an estimate of the RKHS norm might lead to safety constraint violations as we would be too optimistic.

To get a more reliable estimate of~$\|h(\cdot,i)\|_k$, we construct a predictor~$\eta$, which is described in Algorithm~\ref{alg:predictor_training}. 
In particular, we train a multilayer perceptron (MLP) that predicts the RKHS norm~$\|h(\cdot,i)\|_k$ based on the RKHS norm of the mean~$\|\mu_A(\cdot, i)\|_k$ and the reciprocal covariance integral~$r_{\domain,A}$.
Clearly, we require data to train the MLP.
Therefore, we let \safeopt\ optimize random, artificial RKHS functions~$\rho_j(\cdot, i),\, j=1,\ldots, q$, with known RKHS norms~$\|\rho_j(\cdot, i)\|_k$.
%
In each iteration of \safeopt, we compute~$\|\mu_A(\cdot,i)\|_k$ and~$r_{\domain,A}$ and store them as sequences~$X_j$.
Those sequences serve as the inputs for training the MLP, whereas the true RKHS norms of the random functions,~$\|\rho_j(\cdot,i)\|_k$, are the labels.\footnotemark[1]\footnotetext[1]{%
One can, \eg also use $2\cdot \|\rho_j(\cdot, i)\|_k$ as the label to introduce some conservatism.
}
As the sequences~$X_j$ of collected~$\|\mu_A(\cdot,i)\|_k$ and~$r_{\domain,A}$ naturally have different lengths and MLPs work with fixed input sizes, we use zero-padding at the beginning of the inputs.
For evaluating the MLP, we compute the sequence~$X$ containing~$\|\mu_A(\cdot,i)\|_k$ and~$r_{\domain,A}$ up until the current parameter set~$A$ and receive
\begin{align}\label{eq:NN}
    B_i = \eta( X).
\end{align}
After training, the returned value $B_i$ should be a reasonable estimate of~$\|h(\cdot, i)\|_k$.
However, we require an \emph{over}-estimation of~$\|h(\cdot,i)\|_k$ with statistical guarantees. 
Thus, in Section~\ref{sec:theory}, we will add a safety factor to~\eqref{eq:NN}, and analyze the theoretical properties of the algorithm.

\begin{algorithm}
\begin{algorithmic}[1]
\Require RKHS functions~$\rho_j(\cdot,i)$, parameter space~$\domain$, kernel~$k$, MLP architecture
\State Init: $(\overline X, \overline Y)=\emptyset$
\For{each~$\rho_j(\cdot,i)$}
\State Randomly initialize set of safe samples~$S_0$ and initialize~$ X_j = \emptyset$
\For{each \safeopt\ step}
\State Compute~$\|\mu_A(\cdot,i)\|_k$ and~$r_{\domain,A}$ given samples~$A$ and domain~$\domain$ \Comment{GP}
\State $X_j \gets X_j \cup \{\|\mu_A(\cdot,i)\|_k, r_{\domain,A}\}$
\EndFor
\State $\overline X \gets \overline X \cup X_j, \overline Y \gets \overline Y \cup \{\|\rho_j(\cdot,i)\|_k\}$ \Comment{Collect training data for all~$\rho_j(\cdot,i)$}
\EndFor
\State Train MLP with given architecture, inputs~$\overline X$, and labels~$\overline Y$
\caption{Training of the MLP}
\label{alg:predictor_training}
\end{algorithmic}
\end{algorithm}

\subsection{Local interpretation of the RKHS norm} \label{sec:locality}

We have seen that the RKHS norm of the mean function~$\|\mu_A(\cdot, i)\|_k$ is a reasonable estimate of the unknown RKHS function~$\|h(\cdot,i)\|_k$ for a high sampling density~$r_{\domain,A}$.
However, to guarantee safety, we restrict ourselves to a sub-space~$S_A\subseteq \domain$ and consciously leave potentially large parts of the parameter space~$\domain\setminus S_A$ unexplored.
Thus, we can expect a small sampling density~$r_{\domain,A}$ with respect to the complete parameter space~$\domain$ even after many iterations.
At the same time, since we only explore a sub-space, considering the RKHS norm on its complement is superfluous.
Furthermore, it is especially in unsafe regions~$\domain \setminus S_A$ where we might experience non-smooth behavior, \ie large RKHS norms, \eg if the dynamics in those regions are unstable.
Finally, we can exploit local smoothness for more aggressive exploration due to tighter confidence intervals using local RKHS norms.
These arguments motivate a local interpretation of the RKHS norm on sub-domains instead of one RKHS norm for the complete parameter space~$\domain$.

Akin to \safeopt\ and other safe BO algorithms, we start exploration around the set of initial safe samples~$S_0$ and aim to enlarge the safe set progressively.
Hence, a reasonable interpretation of locality is a neighborhood around the samples~$A$.
Therefore, we introduce~$\widetilde \domain$---the convex hull of samples~$A$---which we further expect to be a sub-domain with relatively high sampling density.
We start by restricting exploration to the convex hull~$\widetilde \domain$, with corresponding GPs and RKHS norm upper bounds  $\widetilde{\mathcal{GP}}_i$ and~$\widetilde B_i$, respectively.
To allow for exploration outside of the convex hull~$\widetilde\domain$ while still ensuring a high sampling density, we define a second sub-domain~$\widehat \domain$ with~$\widehat B_i$ and~$\widehat{\mathcal{GP}}_i$ such that~$\widetilde\domain\subseteq\widehat \domain\subseteq \domain$.
Note that we additionally preserve the global RKHS norm estimate~$B_i$ and GPs~$\mathcal{GP}_i$, which yields three total partitions.

\begin{remark}
    Using three partitions, namely~$\widetilde\domain$, $\widehat\domain$, and~$\domain$, is a very simple heuristic, and extending this to more local objects or even an adaptive partitioning of the parameter space is a promising area for further research.
    We want to highlight that locality does not impact safety and mainly improves exploration.
\end{remark}

\subsection{PACSBO}\label{sec:algorithm}
We now combine the RKHS norm estimation and the local interpretation of the RKHS norm into \pacsbo, summarized in Algorithm~\ref{alg:pbo}.
We start by computing the sub-domains~$\widetilde \domain$ and~$\widehat\domain$ (l.~3) and defining the GPs~$\widetilde{\mathcal{GP}}_i$,~$\widehat{\mathcal{GP}}_i$, and~$\mathcal {GP}_i$
corresponding to~$\widetilde\domain$, $\widehat\domain$, and~$\domain$  with samples~$A$, respectively (l.~4).
We next require \emph{guaranteed} upper bounds on the corresponding RKHS norms.
To this end, we query Algorithm~\ref{alg:PAC}, which we present in Section~\ref{sec:theory}. 
Then, we execute the \safeopt-subroutine with each local object and compute the sets of potential maximizers and potential expanders (l.~6).
Finally, we choose the policy parameter to be tried next as the most uncertain policy parameter of the three candidates returned by each \safeopt-subroutine (l.~7) before conducting a new experiment (l.~8), and updating the sample sets (l.~9).
We continue with the described procedure for~$\overline N$ iterations and return our estimate of the optimal safe policy parameter (l.~10).

\begin{algorithm}
\begin{algorithmic}[1]
\Require $k$, $\domain$, $S_0$, $\eta$, $\delta$, $\underline p$, $\overline p$, $\overline N$, $F_\mathrm{safety}$
\State Init: $A\gets S_0$, $\widehat h_{A,i} \gets \widehat  h(S_0,i)$
\For{$1:\overline N$} 
\State  Determine $\widetilde \domain$, $\widehat \domain$ given~$\domain$ and $A$ \Comment{Compute sub-domains}
\State
$\widetilde{\mathcal{GP}}_i, \widehat{\mathcal{GP}}_i, \mathcal{GP}_i\gets$ Define GPs on $\widetilde \domain, \widehat\domain, \domain$ with samples~$A$ 
\State 
$\widetilde B_i$, $\widehat B_i$, $B_i$ $\gets$ Compute RKHS norm over-estimation \Comment{Algorithm~\ref{alg:PAC}}
\State 
$\widetilde M_{ A} \cup \widetilde G_{ A}, \widehat M_{ A} \cup \widehat G_{ A}, M_A \cup G_A \gets $  
\Statex \phantom{\textbf{for}  }\safeopt-subroutines with $(\widetilde B_i, \widetilde{\mathcal{GP}}_i)$, $(\widehat{B}_i, \widehat{\mathcal{GP}}_i)$, $(B_i, \mathcal{GP}_i)$
\State $a\gets$ Most uncertain parameter~$a \in (\widetilde M_{ A} \cup \widetilde G_{ A}) \cup (\widehat M_{ A} \cup \widehat G_{ A}) \cup (M_A \cup G_A) $ 
\State  $\widehat h(a, i)\gets$ Conduct experiment with policy parameter  $a$
\State Append $a$ to $A$, $\widehat h(a, i)$ to $\widehat h_{A,i}$ \Comment{Update sample sets}
\EndFor
\State \Return Best safely evaluable policy parameter
\caption{\pacsbo}
\label{alg:pbo}
\end{algorithmic}
\end{algorithm}

\subsection{Theoretical analysis}\label{sec:theory}
In Section~\ref{sec:RKHS_norm_approx}, we introduced the MLP~\eqref{eq:NN} that returns an estimate of the unknown RKHS norm.
In the following, we propose an algorithm to ensure that~$B_i$ is indeed a probably approximately correct (PAC)~\cite{Shwartz2014Understanding} \emph{over}-estimation of the unknown RKHS norm.
Although we present this section considering the RKHS norm on the parameter space~$\domain$, adjustments to the RKHS norm of the function restricted to the sub-domains~$\widetilde\domain$ or~$\widehat\domain$ work analogously.

The PAC RKHS norm over-estimation is based on  creating~$q$ independent and identically distributed (i.i.d.) random RKHS functions given samples~$A$, $\rho_{A,j}(\cdot,i)\in H_k$, $j\in\{1,\ldots,q\}$, with known RKHS norms~$\|\rho_{A,j}(\cdot,i)\|_k$, similar to the random RKHS functions used for training the MLP.
However, these random RKHS functions are designed to interpolate the given samples subject to $\sigma$-sub-Gaussian noise, and their remaining coefficients are sampled from a bounded interval.
To provide PAC bounds for the RKHS norm over-estimation using random RKHS functions, we require the following assumption. 

\begin{assumption}\label{asm:expectation}
Consider any~$i \in \mathcal I$ and any~$A\subseteq \domain$.
  Given samples $A$, $\|h(\cdot, i)\|_k \leq \lim_{q\rightarrow \infty} \frac{1}{q}\sum_{j=1}^q \|\rho_{A,j}(\cdot,i)\|_k$.
\end{assumption}
Assumption~\ref{asm:expectation} states that the expected value of the RKHS norm of the random functions overestimates the RKHS norm of~$h(\cdot, i)$.
We discuss Assumption~\ref{asm:expectation} in Section~\ref{sec:implementation} as it also depends on the random RKHS function generation.
Finally, we present our main theoretical contribution.
\begin{theorem}\label{th:hoeffding}
Consider any~$i \in \mathcal I$ and let Assumption~\ref{asm:expectation} hold. 
If
\begin{align*}
B_i\geq \frac{1}{q}\sum_{j=1}^q\|\rho_{A,j}\|_k+w_i, \, w_i\coloneqq \sqrt{\frac{\ln(\frac{2}{\delta})(\underset{\alpha,x}{\max} \|\rho_{A}(\cdot, i)\|_k-\underset{\alpha,x}{\min} \|\rho_{A}(\cdot, i)\|_k)^2}{2q}},
\end{align*}
then~$B_i\geq \|h(\cdot, i)\|_k$ with a probability of at least~$1-\delta$, where~$\delta\in(0,1)$.
\end{theorem}
\begin{proof}[Idea]
We use Hoeffding's inequality~\cite[Lemma~4.5]{Shwartz2014Understanding} to upper-bound the deviation between the empirical mean computed by~$q$ random RKHS functions and the true mean computed with infinitely many random RKHS functions.
Then, we leverage Assumption~\ref{asm:expectation} to connect the true mean and the ground truth RKHS norm~$\|h(\cdot,i)\|_k$.
The details can be found in Section~\ref{app:proof}.
\end{proof}

We summarize the PAC RKHS norm over-estimation in Algorithm~\ref{alg:PAC}.
After querying the MLP~\eqref{eq:NN}, we compute~$q$ random RKHS functions and check whether~$B_i$ satisfies the inequality from Theorem~\ref{th:hoeffding} (l.~7).
If~$B_i$ is large enough, we return the RKHS norm over-estimation (l.~12). 
Otherwise, we create more random RKHS functions until a maximum of~$\overline q$ random RKHS functions.
Then, we increase~$B_i$ by a user-chosen safety factor~$F_\mathrm{safety}$ (l.~11).

%

\begin{algorithm}
\begin{algorithmic}[1]
\Require $\delta$, $\underline q$, $\overline q$, $A$, $\domain$, $\eta$, $\widehat h_{A,i}$, $F_\mathrm{safety}$, $k$, $\mathcal{GP}$, $X$
\State Init: $p\gets \underline p$
\State Compute~$\|\mu_{A}(\cdot,i)\|_k$ and~$r_{\domain, A}$ with~$\domain, A,  \widehat h_{A,i}, k$, and~$\mathcal{GP}$ 
\State $X \gets X \cup \{\|\mu_{A}(\cdot,i)\|_k ,r_{\domain,A}\}$
\State $B_i \gets \eta(X)$ \Comment{\eqref{eq:NN}}
\While{True}
\State Compute~$q$ random RKHS functions~$\rho_{A,j}(\cdot, i)$ and~$\|\rho_{A,j}(\cdot,i)\|_k$
\Comment{Section~\ref{sec:implementation}}
\If{
$B_i\geq \frac{1}{q}\sum_{j=1}^q\|\rho_{A,j}(\cdot,i)\|_k + w_i$
} \Comment{Theorem~\ref{th:hoeffding}}
\State \textbf{break}
\EndIf
\State $q \gets q + \underline q$ \Comment{Increase the number of random RKHS functions}
\If{$q > \overline q$}
\State $B_i \gets F_\mathrm{safety} \cdot B_i$ \Comment{Increase~$B_i$ by a safety factor $F_\mathrm{safety}$}
\EndIf
\EndWhile
\State \Return $B_i$
\caption{PAC RKHS norm over-estimation}
\label{alg:PAC}
\end{algorithmic}
\end{algorithm}

%% file: final_figs/RKHS_norm_variance.tex
\begin{tikzpicture}

\definecolor{darkgray176}{RGB}{176,176,176}
\pgfplotsset{
every axis legend/.append style={
at={(1, 0)},
anchor=south east,
},
}
\begin{axis}[
tick align=outside,
tick pos=left,
x grid style={darkgray176},
xlabel={$r_{\domain,A}$},
xmin=1.02085395262562, xmax=8.55259032843921,
xtick style={color=black},
y grid style={darkgray176},
ylabel={RKHS norm},
width=5cm,
height=5cm,
ymin=0.628295874592214, ymax=1.08544943692063,
ytick style={color=black},
legend style={nodes={scale=0.8}}
]
\addplot [semithick, blue]
table {%
1.3632056060717 1
1.41237392123492 1
1.61842873147227 1
1.90802737418646 1
2.02947997092028 1
2.35459887712859 1
2.55298055722506 1
2.78859645139897 1
3.02131840749174 1
3.07769733281788 1
3.1628734552084 1
3.16312130588294 1
3.24675634005761 1
3.32760008151878 1
3.7201013797747 1
3.85529853885701 1
3.9316321220834 1
4.15831323447481 1
4.2430232747333 1
4.29458503068912 1
4.39750096494244 1
4.44824541369619 1
4.58428059250092 1
4.59699059143527 1
5.07195743328702 1
5.07734055163302 1
5.32885963432271 1
5.38827197485267 1
6.4667047616749 1
6.67933790372561 1
6.87415674307701 1
7.10254918510832 1
7.17257204047492 1
7.21267619230325 1
7.76995973567412 1
8.11828198651901 1
8.21023867499314 1
};
\addplot [semithick, black, mark=asterisk, mark size=3, mark options={solid}, only marks]
table {%
1.3632056060717 0.649075581970779
1.41237392123492 0.649920271964394
1.61842873147227 0.674045986035287
1.90802737418646 0.750293167531534
2.02947997092028 0.750665587006885
2.35459887712859 0.766660651630284
2.55298055722506 0.769361404542093
2.78859645139897 0.775453026367097
3.02131840749174 0.808827563789299
3.07769733281788 0.808969224051935
3.1628734552084 0.809591641021081
3.16312130588294 0.809912434386305
3.24675634005761 0.818460347119275
3.32760008151878 0.848363307523837
3.7201013797747 0.859538015631319
3.85529853885701 0.863090287155564
3.9316321220834 0.872050417270678
4.15831323447481 0.875606523238181
4.2430232747333 0.872966029157798
4.29458503068912 0.878646271624018
4.39750096494244 0.882620270446585
4.44824541369619 0.883755699069483
4.58428059250092 0.887238258806654
4.59699059143527 0.887847438920893
5.07195743328702 0.933252859795473
5.07734055163302 0.933460621836885
5.32885963432271 0.93343042481561
5.38827197485267 0.933454225210206
6.4667047616749 0.939745964835553
6.67933790372561 0.939560876715156
6.87415674307701 0.939341005224283
7.10254918510832 0.93969316213956
7.17257204047492 0.939686864557686
7.21267619230325 0.939961406449658
7.76995973567412 0.943364450644862
8.11828198651901 0.947496687864182
8.21023867499314 0.94765931288076
};
\legend{
$\|h{(\cdot, i)}\|_k$,$\| \mu_A{(\cdot, i)}\|_k$
}
\end{axis}

\end{tikzpicture}

%% file: Sections/approximation.tex
\section{Implementation}\label{sec:implementation}
In this section, we discuss the practical implementation of \pacsbo, particularly the random RKHS function generation, and comment on Assumption~\ref{asm:expectation}.

\fakepar{Random RKHS function generation} We need random RKHS functions to obtain training data for the MLP in Algorithm~\ref{alg:predictor_training} and for the PAC validation in Algorithm~\ref{alg:PAC}.
Constructing random RKHS functions considering the entire RKHS~$H_k$ would require us to compute infinite sums, which is impossible.
Hence, we compute functions~$\rho_{A,j}=\sum_{s=1}^{\hat N} \alpha_s k(x_s,\cdot)$ with finite~$\hat N \gg N$, essentially restricting~$\rho_{A,j}(\cdot,i)$ to a pre-RKHS~$H_{0,k}\subseteq H_k$ \cite[Section~2.3]{Kanagawa2018Gaussian}, similar to the pre-RKHS approach presented in~\cite[Appendix~C.1]{Fiedler2021Practical} with random~$\alpha$ and~$x$.
To ensure the interpolating property of the random RKHS functions for the PAC validation, the coefficients~$\alpha_1,\ldots,\alpha_N$ are determined by the samples~$A$ and~$\hat h_{A,i}$.
The remaining coefficients are sampled uniformly within the interval $[-\overline \alpha, \overline \alpha]$.
Moreover, we enforce that the first~$N$ center points of~$\rho_{A,j}(\cdot,i)$ in Algorithm~\ref{alg:PAC} are the samples~$A$ and sample the remaining center points uniformly from~$\domain$.

\begin{figure}
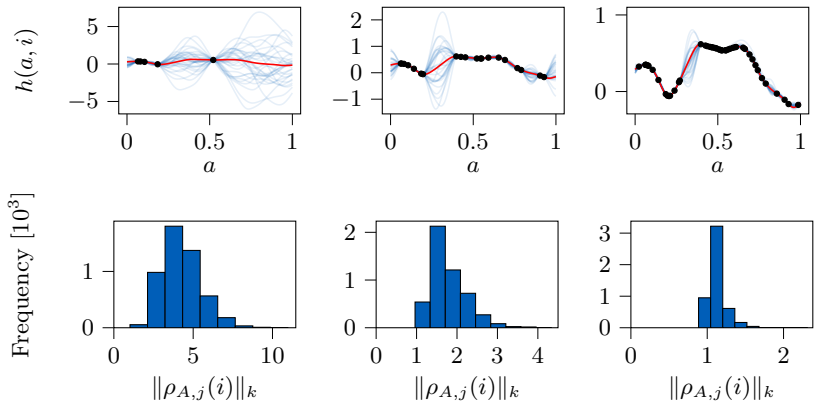

\centering
\input{final_figs/New_Intro_PAC/PAC_functions_5}
\input{final_figs/New_Intro_PAC/PAC_functions_20}
\input{final_figs/New_Intro_PAC/PAC_functions_50}
\hspace*{-0.1cm}
\input{final_figs/New_Intro_PAC/PAC_norms_5}
\hspace*{0.2cm}
\input{final_figs/New_Intro_PAC/PAC_norms_20}
\hspace*{0.1cm}
\input{final_figs/New_Intro_PAC/PAC_norms_50}
\caption{%
Random RKHS functions and their RKHS norms.
\capt{
The upper sub-figures show the ground truth~$h(\cdot,i)$ (red line) and the random RKHS functions~$\rho_{A,j}(\cdot,i)$ (blue lines) given samples~$A$ (black dots), while the lower sub-figures illustrate the frequency of the RKHS norms~$\|\rho_{A,j}(\cdot,i)\|_k$. 
For an increasing sample set, $\|\rho_{A,j}(\cdot,i)\|_k$ approach~$\|h(\cdot,i)\|_k=1$, whereas~$\|\rho_{A,j}(\cdot,i)\|$ tends to conservatively over-estimate~$\|h(\cdot,i)\|_k$ for fewer samples.
We used the Matérn-32 kernel with length scale~$\ell=0.1$, $\hat N=100$, and~$\overline\alpha=1$.
We sampled the parameters~$A$ randomly from a uniform distribution.
The upper sub-figures show a random subset of 30 out of the~$q=5\cdot 10^3$ random RKHS functions.}
}
\label{fig:intro_PAC}
\end{figure}

\fakepar{Assumption~\ref{asm:expectation}, tightness, and randomness}
To account for Assumption~\ref{asm:expectation}, \ie to ensure that the RKHS norm of the random RKHS functions over-estimate the unknown RKHS norm in expectation, we choose~$\overline\alpha$ and~$\hat N$ such that $\|\rho_{A,j}(\cdot,i)\|_k$ can get large for small sample sets~$A$, which is illustrated in Figure~\ref{fig:intro_PAC}.
Figure~\ref{fig:intro_PAC} also shows that~$\|\rho_{A,j}(\cdot,i)\|_k$  approaches~$\|h(\cdot,i)\|_k$ for an increasing sample set~$A$.
Hence, we expect Algorithm~\ref{alg:PAC} to return a tighter over-estimate~$B_i\geq \|h(\cdot,i)\|_k$ for more samples~$A$.
In particular, for the three toy examples in Figure~\ref{fig:intro_PAC} with~$\|h(\cdot,i)\|_k=1$ and~$q=5\cdot 10^3$, Algorithm~\ref{alg:PAC} terminates whenever~$B_i$ is greater than $5.71$ (left, 5 samples), $2.96$ (center, 20 samples), and $1.71$ (right, 50 samples), respectively, which supports that we can obtain tighter RKHS norm over-estimations with denser sampling. 

\begin{remark}
Our design choices reduce the randomness of the random RKHS functions.
Furthermore, as mentioned in~\cite[Appendix~C.1]{Fiedler2021Practical}, generating RKHS functions by following the pre-RKHS approach is prone to introducing a slight bias.
Nevertheless, our numerical investigations display adequate randomness and capture a wide range of RKHS functions. 
\end{remark}

\begin{remark}
It is possible to design RKHS functions~$h(\cdot,i)$ for which Assumption~\ref{asm:expectation} is violated, and hence Theorem~\ref{th:hoeffding} becomes inapplicable. 
This can especially happen with a very large ground truth RKHS norm combined with a small number of samples and a too small~$\overline\alpha.$
The former may cause a relatively small mean RKHS norm~$\|\mu_A(\cdot,i)\|_k$, which may result in
a too small prediction of the MLP. 
We reduce this risk by following a localized approach and increasing sampling density.
Furthermore, we choose a large enough bound on the coefficients~$\overline\alpha$ to capture large ground truth RKHS norms.
Finally, the toy example in Figure~\ref{fig:intro_PAC} and the forthcoming experiments in Section~\ref{sec:result} support the sensibility and applicability of our approach.
Notably, even if Assumption~\ref{asm:expectation} is violated, we still leverage collected samples and a sophisticated predictor to estimate the RKHS norm.
In contrast, \safeopt\ guesses the RKHS norm bound a priori.
Particularly, numerous \safeopt-applications, \eg\cite{Berkenkamp2021Bayesian}, use a constant multiple of the posterior covariance as confidence intervals when conducting experiments.
\end{remark}

%% file: final_figs/New_Intro_PAC/PAC_norms_5.tex
\begin{tikzpicture}

\definecolor{darkgray176}{RGB}{176,176,176}

\begin{axis}[
tick align=outside,
tick pos=left,
x grid style={darkgray176},
xlabel={$\|\rho_{A,j}(i)\|_k$},
xmin=0,
xmax=11.4684303984046,
xtick style={color=black},
y grid style={darkgray176},
ylabel={ {Frequency $[10^3]$}},
ymin=0, ymax=1893.15/1000,
ytick style={color=black},
tick label style={font=\small},
ytick={0,1},
width=4cm,
height=3cm,
]
\draw[draw=black,fill=aaltoBlue] (axis cs:1.01687875986099,0) rectangle (axis cs:2.12286306023598,54/1000);
\draw[draw=black,fill=aaltoBlue] (axis cs:2.12286306023598,0) rectangle (axis cs:3.22884736061096,983/1000);
\draw[draw=black,fill=aaltoBlue] (axis cs:3.22884736061096,0) rectangle (axis cs:4.33483166098595,1803/1000);
\draw[draw=black,fill=aaltoBlue] (axis cs:4.33483166098595,0) rectangle (axis cs:5.44081596136093,1373/1000);
\draw[draw=black,fill=aaltoBlue] (axis cs:5.44081596136093,0) rectangle (axis cs:6.54680026173592,565/1000);
\draw[draw=black,fill=aaltoBlue] (axis cs:6.54680026173592,0) rectangle (axis cs:7.6527845621109,179/1000);
\draw[draw=black,fill=aaltoBlue] (axis cs:7.6527845621109,0) rectangle (axis cs:8.75876886248589,33/1000);
\draw[draw=black,fill=aaltoBlue] (axis cs:8.75876886248589,0) rectangle (axis cs:9.86475316286087,8/1000);
\draw[draw=black,fill=aaltoBlue] (axis cs:9.86475316286087,0) rectangle (axis cs:10.9707374632359,1/1000);
\end{axis}

\end{tikzpicture}

%% file: final_figs/New_Intro_PAC/PAC_norms_20.tex
\begin{tikzpicture}

\definecolor{darkgray176}{RGB}{176,176,176}

\begin{axis}[
tick align=outside,
tick pos=left,
x grid style={darkgray176},
xlabel={$\|\rho_{A,j}(i)\|_k$},
xmin=0,
xmax=4.49957287055254,
xtick style={color=black},
y grid style={darkgray176},
ymin=0, ymax=2234.4/1000,
ytick style={color=black},
width=4cm,
tick label style={font=\small},
ytick={0, 1, 2},
height=3cm
]
\draw[draw=black,fill=aaltoBlue] (axis cs:0.962220275402069,0) rectangle (axis cs:1.33654330134392,539/1000);
\draw[draw=black,fill=aaltoBlue] (axis cs:1.33654330134392,0) rectangle (axis cs:1.71086632728577,2128/1000);
\draw[draw=black,fill=aaltoBlue] (axis cs:1.71086632728577,0) rectangle (axis cs:2.08518935322762,1211/1000);
\draw[draw=black,fill=aaltoBlue] (axis cs:2.08518935322762,0) rectangle (axis cs:2.45951237916946,722/1000);
\draw[draw=black,fill=aaltoBlue] (axis cs:2.45951237916946,0) rectangle (axis cs:2.83383540511131,271/1000);
\draw[draw=black,fill=aaltoBlue] (axis cs:2.83383540511131,0) rectangle (axis cs:3.20815843105316,85/1000);
\draw[draw=black,fill=aaltoBlue] (axis cs:3.20815843105316,0) rectangle (axis cs:3.58248145699501,24/1000);
\draw[draw=black,fill=aaltoBlue] (axis cs:3.58248145699501,0) rectangle (axis cs:3.95680448293686,16/1000);
\draw[draw=black,fill=aaltoBlue] (axis cs:3.95680448293686,0) rectangle (axis cs:4.33112750887871,4/1000);
\end{axis}

\end{tikzpicture}

%% file: final_figs/New_Intro_PAC/PAC_norms_50.tex
\begin{tikzpicture}

\definecolor{darkgray176}{RGB}{176,176,176}

\begin{axis}[
tick align=outside,
tick pos=left,
x grid style={darkgray176},
xlabel={$\|\rho_{A,j}(i)\|_k$},
xmin=0,
xmax=2.38346838784218,
xtick style={color=black},
y grid style={darkgray176},
ymin=0, ymax=3379.95/1000,
ytick style={color=black},
width=4cm,
height=3cm,
tick label style={font=\small}
]
\draw[draw=black,fill=aaltoBlue] (axis cs:0.886491537094116,0) rectangle (axis cs:1.04490178585052,949/1000);
\draw[draw=black,fill=aaltoBlue] (axis cs:1.04490178585052,0) rectangle (axis cs:1.20331203460693,3219/1000);
\draw[draw=black,fill=aaltoBlue] (axis cs:1.20331203460693,0) rectangle (axis cs:1.36172228336334,612/1000);
\draw[draw=black,fill=aaltoBlue] (axis cs:1.36172228336334,0) rectangle (axis cs:1.52013253211975,167/1000);
\draw[draw=black,fill=aaltoBlue] (axis cs:1.52013253211975,0) rectangle (axis cs:1.67854278087616,42/1000);
\draw[draw=black,fill=aaltoBlue] (axis cs:1.67854278087616,0) rectangle (axis cs:1.83695302963257,6/1000);
\draw[draw=black,fill=aaltoBlue] (axis cs:1.83695302963257,0) rectangle (axis cs:1.99536327838898,2/1000);
\draw[draw=black,fill=aaltoBlue] (axis cs:1.99536327838898,0) rectangle (axis cs:2.15377352714539,2/1000);
\draw[draw=black,fill=aaltoBlue] (axis cs:2.15377352714539,0) rectangle (axis cs:2.31218377590179,1/1000);
\end{axis}

\end{tikzpicture}

%% file: Sections/experiments.tex
\section{Results} \label{sec:result}
In this section, we present numerical experiments and a hardware experiment using \pacsbo.
All experiments used the Matérn-32 kernel with length scale~$\ell=0.1$ and confidence~$1-\delta=0.9$.
Moreover, we choose the sub-domain~$\widehat \domain$
to be the \SI{10}{\percent} uniform enlargement of the convex hull~$\widetilde \domain$, and set~$\domain=[0,1]^n$.

\subsection{Numerical experiments} \label{sec:numerical}
\fakepar{Setting}
We compare \pacsbo\ and \safeopt\ by maximizing an unknown RKHS function~$h(\cdot,0)$ with~$\|h(\cdot,0)\|_k=2$.
We have~$\Ig=\{1\}$ and set~$h(\cdot,1)=h(\cdot,0)-f_{\mathrm g}$, where~$f_\mathrm{g}\in\mathbb{R}$ is the pre-defined safety threshold.
\pacsbo\ and \safeopt\ both start with the same three policy parameters~$S_0$.

\begin{figure}
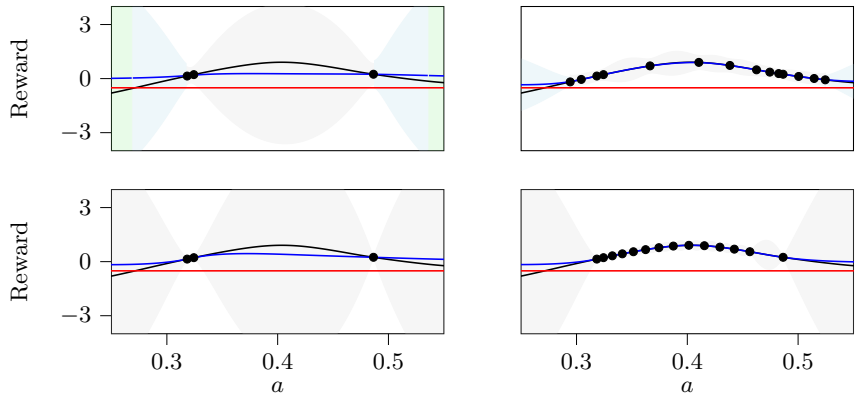

\centering
\begin{subfigure}{6.25cm}
    \centering
    \input{final_figs/num_exp/too_large_10/PACSBO_iteration_3}
\end{subfigure}
\begin{subfigure}{5.5cm}
    \centering
\input{final_figs/num_exp/too_large_10/PACSBO_iteration_14}   
\end{subfigure}
\begin{subfigure}{6.25cm}
    \centering
\input{final_figs/num_exp/too_large_10/SafeOpt_iteration_3}
\end{subfigure}
\begin{subfigure}{5.5cm}
    \centering
\input{final_figs/num_exp/too_large_10/SafeOpt_iteration_14}
\end{subfigure}
\caption{\pacsbo\ vs.\ \safeopt. 
\capt{
A conservative smoothness assumption allows \pacsbo\ (upper figures)  to explore faster than \safeopt\ (lower figures).
}
}
\label{fig:large}
\end{figure}

\fakepar{Results}
Figure~\ref{fig:large} compares \pacsbo\ with~$\hat N=100$ and~$\overline\alpha=1$ and \safeopt\ with~$B_i=10$. 
Note that this choice of~$\hat N$ and~$\overline\alpha$ 
heuristically 
yields random RKHS functions with RKHS norms of up to 10 (see Figure~\ref{fig:intro_PAC}).
Depicted are the GP mean (blue line), the reward~$h(\cdot,0)$ (black line), the samples~$A$ (black dots), and the safety threshold~$f_\mathrm{g}$ (red line).
The confidence intervals of \pacsbo\ for $\tilde\domain$, $\hat\domain$, and $\domain$ are depicted in grey, blue, and green, respectively, whereas the confidence intervals of \safeopt\ are shown in grey.
\pacsbo\ shows superior exploration compared to \safeopt, mainly because: \emph{(i)} \pacsbo\ learns the RKHS norm from data and is, thus, less conservative; \emph{(ii)} \pacsbo\ makes use of a local interpretation of the RKHS norm.

Figure~\ref{fig:small} compares \pacsbo\ with~$\hat N=100$ and~$\overline\alpha=0.04$ and \safeopt\ with $B=0.4$ for a function~$h(\cdot,0)$ with~$\|h(\cdot,0)\|_k=2$, \ie both algorithms over-estimate the smoothness of~$h(\cdot,0)$.
\safeopt\ cannot correct that assumption and samples unsafely.
In contrast, \pacsbo\ learns the RKHS norm and stays safe.
Note that a too small choice of~$\overline\alpha$ might lead to Algorithm~\ref{alg:PAC} returning the estimate~\eqref{eq:NN} of the predictor without a safety factor.
Hence, an accurate predictor~\eqref{eq:NN} can still return a reasonable estimate of the RKHS norm.
However, the obtained PAC bounds would become mundane.


\begin{figure}
\centering
\begin{subfigure}{6.25cm}
    \centering
    \input{final_figs/num_exp/too_small_0.4/PACSBO_iteration_3}
\end{subfigure}
\begin{subfigure}{5.5cm}
    \centering
\input{final_figs/num_exp/too_small_0.4/PACSBO_iteration_14}   
\end{subfigure}
\begin{subfigure}{6.25cm}
    \centering
\input{final_figs/num_exp/too_small_0.4/SafeOpt_iteration_3}
\end{subfigure}
\begin{subfigure}{5.5cm}
    \centering
\input{final_figs/num_exp/too_small_0.4/SafeOpt_iteration_14}
\end{subfigure}
\caption{\pacsbo\ vs.\ \safeopt.
\capt{
An optimistic smoothness assumption yields unsafe experiments (red cross) with \safeopt\ (bottom), whereas \pacsbo\ (top) stays safe.
}
} 
\label{fig:small}
 \end{figure}


\subsection{Hardware experiment} \label{sec:hardware}
We next evaluate \pacsbo\ on a real Furuta pendulum~\cite{furuta1992swing}.

\begin{wrapfigure}[12]{r}{0.50\textwidth}
\vspace*{-20pt}
\centering
\input{final_figs/hardware_experiment.tex}
\vspace*{-10pt}
    \caption{
    \pacsbo\ for Furuta pendulum.
    }
    \label{fig:hardware}
\end{wrapfigure}
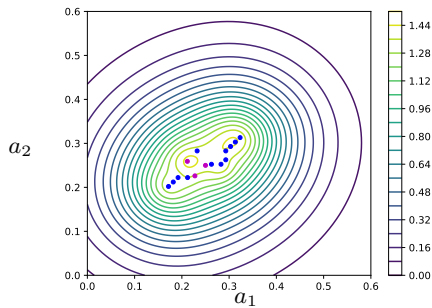
\fakepar{Setting}
We consider the same experimental setup as \cite{Baumann2021GO}.
In particular, we consider state-feedback control to stabilize the pendulum in an upright position and tune the first two entries of the static gain matrix.
We model~$h(\cdot,i)$ with~$\Ig=\{1\}$ as independent GPs.
Particularly, the reward~$h(\cdot,0)$ encourages stable behavior while the constraint~$h(\cdot,1)$ considers the angle between the pendulum and the upright position; see~\cite{Baumann2021GO} for more details.
For \pacsbo, we use~$\hat N=100$ and~$\overline\alpha=1$.

\fakepar{Results}
We show the explored region of~$h(\cdot,1)$ after 15 samples in Figure~\ref{fig:hardware}, where the purple dots illustrate the initial sample set~$S_0$.
We see that \pacsbo\ explores the parameter space without prior knowledge of an RKHS norm upper bound.
Importantly, the algorithm did not incur any failure during exploration.
This demonstrates that \pacsbo\ can safely learn control policies for hardware systems.

%% file: final_figs/hardware_experiment.tex
\begin{tikzpicture}[xscale=1, yscale=1]
\centering
\pgfplotsset{
every axis legend/.append style={
at={(0.5,0.5)},
anchor=north west,
},
}
\node[inner sep=0pt] (whitehead) at (0, 0)
    {\includegraphics[width=5cm]{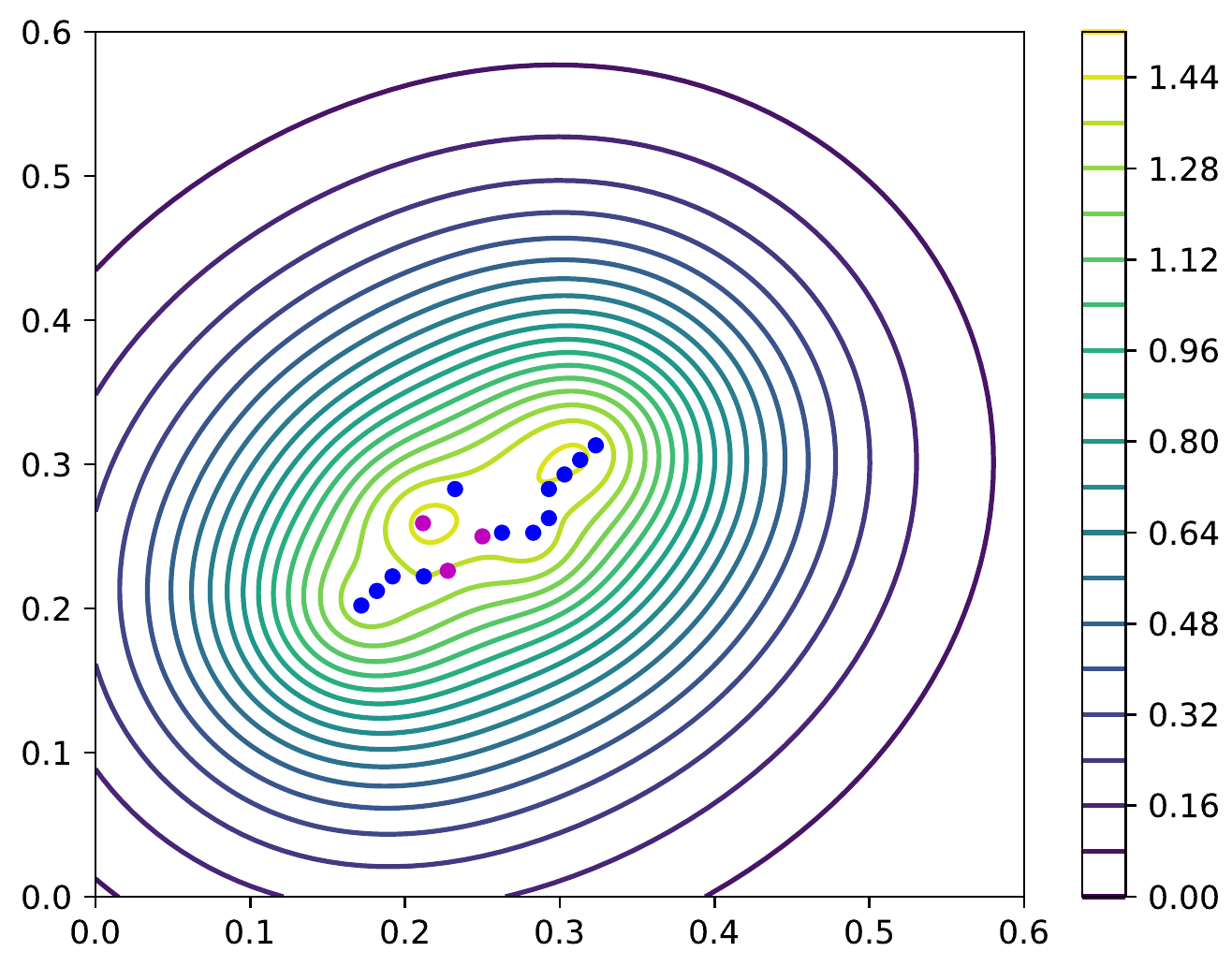}};
\node[align=left] at (-3, 0){{\small $a_2$}};
\node[align=left] at (0, -2){{\small $a_1$}};

\end{tikzpicture}

%% file: Sections/summary.tex
\section{Conclusions}
In this paper, we presented \pacsbo, a novel algorithm for probably approximately correct safe BO.
Unlike prior works on safe BO, especially \safeopt\ and extensions of it, we drop the assumption of knowing tight upper bounds on the RKHS norms of underlying reward and constraint functions.
Instead, we estimate the upper bound from data and theoretically investigate the RKHS norm estimation.
Besides, we treat the RKHS norm as a local object in the area in which we are interested.
This local treatment reduces conservatism compared to assuming one global upper bound on the entire parameter space.
We evaluate \pacsbo\ in numerical experiments and highlight its benefits compared to \safeopt.
Moreover, we demonstrate the successful deployment of \pacsbo\ in a hardware experiment.

This paper is a first step toward removing the RKHS norm upper bound as a standing assumption in safe BO and replacing it with a data-driven estimate.
As such, it leaves room for further work and improvements, starting from a more sophisticated design for the predictor, over different ways of approaching the locality of the RKHS norm, to proving safety and investigating scalability of \pacsbo.
Those extensions and investigations are left for future work.

%% file: Sections/appendix.tex
\section{Proof of Theorem~\ref{th:hoeffding}}\label{app:proof}
We seek to bound the maximum deviation between the empirical average of the norm of our random RKHS functions from their expected value.
To this end, we leverage Hoeffding's inequality~\cite[Lemma~4.5]{Shwartz2014Understanding}.
Thus, we first show that the norms of the random RKHS functions are i.i.d.\ random variables conditioned on the samples.
As we describe in Section~\ref{sec:implementation}, the randomness of the functions comes from the randomness of the coefficients~$\alpha$ and the center points~$x$. 
Specifically, given~$A\subseteq \domain$, $x_{1:N}\coloneqq [x_{1}, \ldots, x_{\mathrm N}]$ and~$\alpha_{1:N}\coloneqq [\alpha_{1},\ldots,\alpha_{\mathrm N}]$ are determined by the interpolating property and we independently sample~~$x_{N{+}1:\hat N}$ and~$\alpha_{N+1{:}\hat N}$ from a uniform distribution.
If the mapping represented by the RKHS norm from the i.i.d.\ coefficients and center points is measurable, then the RKHS norms are i.i.d.\ random variables~\cite[Theorem~1.3.5]{durrett2019probability}.

Choose any~$A\subseteq\domain$ and consider the function $
 F: (\mathbf R^{\hat N-N}, \mathcal R^{\hat N-N}) \rightarrow (\mathbf R_{\geq 0}, \mathcal R_{\geq 0}), F(O)=\sqrt{\sum_{s=1}^{\hat N}\sum_{t=1}^{\hat N} \alpha_s\alpha_t k(x_s,x_t)},
$
which maps from vectors of random center points and coefficients to the RKHS norm with corresponding Borel~$\sigma$-algebras.
We next prove that~$F$ is a measurable function~\cite[Chapter~4.1]{dibenedetto2002real}.
Suppose that~$F$ is not measurable, \ie $\exists E\in\mathcal R_{\geq 0}: \{O \in \mathbf{R}^{\hat N-N} \vert F(O)\in E\} \not \in \mathcal R^{\hat N-N}.$
In words, if~$F$ is not measurable, then there exists an event $E\in\mathcal R_{\geq 0}$
that contains~$F(O)$ of outcomes~$O\in\mathbf{R}^{\hat N-N}$ such that the pre-image of~$E$ under~$F$, \ie the outcomes~$O$ that generated~$F(O)$, are not in the $\sigma$-algebra $\mathcal R^{\hat N-N}$, which is a contradiction since the $\sigma$-algebra is the Borel~$\sigma$-algebra. Hence,~$F$ is measurable.
By exploiting~\cite[Theorem~1.3.5]{durrett2019probability}, we deduce that~$F(O)$ is a random variable and, therefore,  $\|\rho_{A,j}\|_k$ are i.i.d.\ random variables.
Naturally, their expected value is given by~$\lim_{s\rightarrow \infty}\sum_{j=1}^s\frac{1}{s}\|\rho_{A,j}(\cdot, i)\|_k$.
Hence, we can state:
\begin{proposition}\label{le:hoeffding}
Let~$\|\rho_{A,1}\|_k,\ldots, \|\rho_{A,q}\|_k$ be a sequence of i.i.d.\ random variables and choose any $\delta\in(0,1)$.
Then, with probability of at least~$1-\delta$,
it holds that $
\left\lvert \frac{1}{q}\sum_{j=1}^q \|\rho_{A,j}(\cdot,i)\|_k - \lim_{s\rightarrow \infty}\frac{1}{s}\sum_{j=1}^s\|\rho_{A,j}(\cdot, i)\|_k \right\rvert \leq w_i$.
\end{proposition}
\begin{proof}
    The proposition directly follows from Hoeffding's inequality~\cite[Lemma~4.5]{Shwartz2014Understanding}.
\end{proof}
From~Proposition~\ref{le:hoeffding}, we can deduce that
\begin{align}\label{eq:bound_mean}
    \lim_{s\rightarrow \infty}\frac{1}{s}\sum_{j=1}^s\|\rho_{A,j}(\cdot,i)\|_k  \leq \frac{1}{q}\sum_{j=1}^q\|\rho_{A,j}(\cdot,i)\|_k + w_i,
\end{align}
holds with a probability of at least~$1-\delta$.
Therefore, with probability~$1-\delta$,
\begin{align*}
B_i \overset{\mathrm{Alg.}\ref{alg:PAC}}&{\geq}  
\frac{1}{q}\sum_{j=1}^q\|\rho_{A,j}(\cdot,i)\|_k + w_i
\overset{\eqref{eq:bound_mean}}{\geq} \lim_{s\rightarrow \infty}\frac{1}{s}\sum_{i=1}^s\|\rho_{A,i}\|_k 
\overset{\mathrm{Asm.}\ref{asm:expectation}}{\geq} \|h(\cdot,i)\|_k.
\end{align*}